# Artificial Intelligence in Pronunciation Teaching: Use and Beliefs of Foreign Language Teachers

Georgios P. Georgiou[1] [2]

[1]*Department of Languages and Literature, University of Nicosia*

[2]*Director of the Phonetic Lab, University of Nicosia*

## Abstract

Pronunciation instruction in foreign language classrooms has often been an overlooked area of focus. With the widespread adoption of Artificial Intelligence (AI) and its potential benefits, investigating how AI is utilized in pronunciation teaching and understanding the beliefs of teachers about this tool is essential for improving learning outcomes. This study aims to examine how AI use for pronunciation instruction varies across different demographic and professional factors among teachers, and how these factors, including AI use, influence the beliefs of teachers about AI. The study involved 117 English as a Foreign Language (EFL) in-service teachers working in Cyprus, who completed an online survey designed to assess their beliefs about the effectiveness of AI, its drawbacks, and their willingness to integrate AI into their teaching practices. The results revealed that teachers were significantly more likely to agree on the perceived effectiveness of AI and their willingness to adopt it, compared to their concerns about its use. Furthermore, teachers working in higher education and adult education, as well as those who had received more extensive training, reported using AI more frequently in their teaching. Teachers who utilized AI more often expressed stronger agreement with its effectiveness, while those who had received more training were less likely to express concerns about its integration. Given the limited training that many teachers currently receive, these findings demonstrate the need for tailored training sessions that address the specific needs and concerns of educators, ultimately fostering the adoption of AI in pronunciation instruction.

## 1. Introduction

Since its seminal conceptualization in 1956 as the scientific and engineering discipline dedicated to developing intelligent machines (McCarthy, 2007), Artificial Intelligence (AI) has evolved alongside massive technological advancements, including tools that mimic human cognition (Georgiou, 2025a, 2025b). These advancements have inevitably impacted foreign language education, which has undergone rapid transformation in recent years due to the AI revolution.

In the domain of pronunciation, current practice typically includes widely used and cited tools, such as intelligent personal assistants, chatbots, and learning applications (Vančová, 2023), which have the potential to improve learning outcomes. Pronunciation instruction and research have long been overshadowed by other linguistic areas, leading to their marginalization (Georgiou, 2019; Low, 2021). According to Georgiou (2020), pronunciation is important for effective communication and, consequently, the mutual understanding of the speakers. Levis (2022) characterizes it as a socially connected skill, which is unavoidable and essential. Thankfully, advancements in AI provide a valuable opportunity to bring pronunciation instruction to the forefront, making it both more engaging and effective.

While the benefits of using AI tools for pronunciation instruction are clear (Mompean, 2024), the extent of their use may vary, potentially influenced by teachers' prior cognitions (Liu et al., 2024) as well as other characteristics. It is essential to first understand how the individual characteristics of educators affect their teaching practices and how their perceptions are influenced by these characteristics. This study explores teachers' beliefs about AI in the overlooked area of pronunciation instruction, aiming to link these beliefs to various demographic and professional factors. It also examines how AI use for pronunciation teaching varies based on these factors.

## 1.1 Impact of AI on pronunciation instruction

Pronunciation is currently integrated within the broader framework of speaking skills, as speaking and pronunciation are inherently connected (Vančová, 2023). Pronunciation instruction goes beyond merely perfecting individual sounds (segmental level) (e.g., Ghorbani et al., 2016), also encompassing prosodic features, such as intonation, rhythm, and stress (suprasegmental level) (e.g., De-la-Mota, 2019). However, attaining complete accuracy in pronunciation is challenging and often unnecessary for most nonnative language learners (Levis, 2022). Instead, the primary focus of modern pronunciation training is on enhancing comprehensibility and intelligibility in speech (Derwing & Munro, 2015; Pennington, 2021). With the rapid advancement of technology, including AI, and the vast array of innovations available, integrating technology for pronunciation teaching has become necessary for educators (Liu & Reed, 2022).

Several studies have demonstrated the positive outcomes of AI use for the learning of foreign language pronunciation. For example, Qiao and Zhao (2023) investigated the effectiveness of AI-based instruction in improving second language speaking skills and speaking self-regulation in a natural setting. The research was conducted with Chinese students of English as a Foreign Language (EFL), who were randomly assigned to either an experimental group receiving AI-based instruction or a control group receiving traditional instruction. The AI-based instruction utilized the Duolingo application, which incorporated natural language processing technology, interactive exercises, personalized feedback, and speech recognition technology. The findings revealed that the experimental group showed significantly greater improvement in L2 speaking skills compared to the control group and participants in the experimental group reported higher levels of speaking self-regulation. Vančová (2023) explored the existing use of AI-powered tools in foreign language pronunciation training through a meta-analysis of 15 research papers. These studies highlighted the benefits of using AI and AI-based tools (such as mobile and web apps, chatbots, and virtual assistants) and offered suggestions for their future applications in educational settings. The existing use of AI tools confirmed positive outcomes in improving intelligibility, increasing motivation, and addressing speaking anxiety among foreign language learners in both formal and informal learning environments.

In another study, Mohammadkarimi (2024) examined the potential of AI to improve English pronunciation skills among elementary, pre-intermediate, and experienced English teachers. A mixed-method approach was employed, combining quantitative and qualitative data collection to evaluate the effectiveness of AI-based pronunciation tools and assess learners' perceptions. Over a two-month period, the experimental group used AI tools such as Listnr and Murf, while the control group received traditional instruction. Data were gathered from pre- and post-test pronunciation scores, questionnaires, and interviews. The results revealed that the experimental group exhibited significant improvements in pronunciation accuracy. The learners expressed positive attitudes toward AI tools, highlighting their effectiveness in improving pronunciation, increasing confidence, and enhancing engagement. While recent studies highlight the importance of AI in pronunciation instruction, research on its perceived values and drawbacks and the willingness of language educators to adopt it remains scarce.

**1.2 Teachers' beliefs about AI and connection with their teaching practices**

Teachers' beliefs influence their actions throughout the learning process. Hew and Brush (2007) described them as assumptions or perceptions about a subject that are regarded as true. These beliefs have been widely recognized as a crucial aspect of teaching and teacher education, as they shape the actions of educators and decision-making (Cheng et al., 2022) and influence student outcomes (Sabarwal et al., 2022). Borg (2019) argued that teachers' beliefs, along with attitudes, knowledge, emotions, perceptions, and thoughts, are part of teacher cognition, which refers to the personal and hidden aspects of their work; these elements have a significant impact on the teachers' actions and teaching practices. In regard to technology, value beliefs of teachers are among the most widely studied cognitive factors (Liu et al., 2024). Value beliefs about technology reflect the views of teachers on the importance and effectiveness of technology in achieving key educational objectives (Lai et al., 2022), and they are crucial constructs that directly impact the use of technology by teachers in their practices (Cheng et al. 2020). Given the varying conceptualizations of these beliefs, this paper defines them in terms of the perceived benefits, aligning conceptually with what Cheng et al. (2020) describe as utility (usefulness) value. The metanalytical study by Wilson (2023) indicated a direct positive relationship between the utility value beliefs of teachers about technology and their integration of technology. The paper also discusses perceived drawbacks of AI (e.g., Delello et al., 2023) and educators' willingness to adopt it for pronunciation instruction.

AI integration in foreign language education is still in its early stages. While AI technology is generally embraced and supported (An et al., 2023), it also presents several perceived challenges, including ethical and security concerns (Dincer & Bal, 2024) and a lack of preparedness among teachers (Dilzhan, 2024). Further research may provide a better understanding of the relationship between teachers' beliefs about AI and their actual use of it. For instance, Varsamidou (2024) investigated views and beliefs of Greek EFL teachers about AI in education and found that 69% had rarely or never used AI in their teaching, while only 31% incorporated it at least occasionally. However, a striking contrast emerged: 80% of teachers believed AI was at least somewhat effective in enhancing the language and communication skills of students. This discrepancy may be attributed to a lack of AI training and the perceived challenges associated with its implementation.

The case study by Kohnke et al. (2023) shed light on generative AI preparedness among university English language instructors in Hong Kong. The authors reported that although the educators had varying levels of experience with AI tools – some already incorporated them into their courses while others were just starting to explore their potential – they acknowledged the broad applicability of these tools in both language learning and the professional practices of language teachers.

The use of AI, especially in pronunciation teaching, is an overlooked area that deserves more attention in the future, given its vast potential (Vančová, 2023). Research on perceptions of AI in foreign language teaching has largely explored the topic from a broad perspective, without specifically addressing pronunciation (e.g., Varsamidou, 2024). When pronunciation has been the focus, studies have typically approached it from the learners' point of view (e.g., Amin, 2024). A recent study by Pokrivčáková (2023) studied the attitudes of pre-service teachers towards AI and its integration into EFL teaching and learning in Slovakia. An online questionnaire was used to collect data. Teachers expressed their level of agreement on whether AI could assist with repetitive tasks such as pronunciation drills. The findings showed that 34% either strongly disagreed or disagreed, 19% were uncertain, and 47% either agreed or strongly agreed with the statement. Interestingly, the connection between teachers' beliefs and their use of AI together with other characteristics in pronunciation teaching could open up further avenues for understanding how these perceptions and characteristics influence classroom practices, as well as identifying potential barriers or facilitators to more widespread adoption of AI tools in this understudied area.

**1.3 How various factors influence teachers' use of AI and their beliefs about AI**

The effect of teachers' demographic and professional profiles on the use of AI as well as their beliefs warrants further investigation. For instance, it could be expected that older and more experienced teachers might be more hesitant to adopt AI due to their comfort with traditional methods and the stereotype that they possess less technological expertise compared to younger educators (Mariano et al., 2022). Cabero-Almenara et al. (2024) examined teachers' acceptance of educational AI and related it to several variables, including age. The findings demonstrated that age did not account for the extent to which teachers used AI or their intention to use it. However, younger teachers believed that the use of AI does not require much effort and that its use will provide them with some enjoyment and pleasure. Nyabaa et al. (2024) studied pre-service teachers' uses and attitudes towards AI in Ghana. The authors found that background factors, such as age and year of study could predict the frequency of AI use. Nevertheless, the attitudes towards AI could not be predicted by these background factors despite the teachers' diverse backgrounds. By contrast, Georgiou (2019) concluded that older EFL teachers in Cyprus held different cognitions than their younger counterparts, which contributed to the neglect of pronunciation teaching. However, this study did not incorporate the AI element.

Alghamdi (2023) researched, among other aspects, whether EFL teachers' views on the TPACK model, which measures constructs related to teaching pronunciation, varied based on factors such as major, experience, and technological competence. Regarding experience, the results revealed no statistically significant difference, suggesting that the instructors' level of experience did not impact their use of technology in teaching pronunciation; however, both novice and experienced instructors agreed on the importance of incorporating technology into pronunciation instruction. Furthermore, the study of Galindo-Domínguez et al. (2024) investigated whether teacher digital competence of Spanish teachers is linked to teacher attitudes towards AI, and if so, whether factors such as educational stage, age, gender, years of experience, and field of knowledge influence this relationship. The findings indicated that, irrespective of the aforementioned factors, higher digital competence in teachers was associated with more positive attitudes towards AI.

For the adoption of AI, teachers require comprehensive training to develop the necessary digital skills (Nazaretsky et al., 2022). Equipping them with AI-related knowledge and skills is essential, as it can enhance their teaching practices and improve overall instructional quality (De La Higuera, 2019). The primary goal of AI training for teachers is to develop their digital competencies (Lee & Perret, 2022) while addressing the difficulties they encounter when using AI tools; a lack of awareness prevents teachers from implementing AI. When instructors feel confident and competent in using AI, they are more likely to develop a positive attitude and a stronger intention to incorporate it into their teaching (Rahiman & Kodikal, 2024). Molefi et al. (2024) argued that, apart from the perceived effectiveness of AI, additional support, such as professional training, is needed for teachers to integrate AI into their teaching. A key observation made by Aljemely (2024) is that AI use training tends to focus primarily on students, while research addressing the needs of teachers remains underexplored.

While previous studies have investigated the impact of demographic and professional factors on teachers' beliefs and use of AI, the existing body of research has overlooked pronunciation instruction. Therefore, new research focusing on this area could provide valuable insights and a deeper understanding of how these factors may affect perceptions and use of AI in pronunciation teaching.

**1.4 This study**

This study fills a research gap by investigating the use of AI by EFL teachers in Cyprus for pronunciation teaching as well as their beliefs about AI for this purpose. Specifically, it aims to answer the following research questions:

a) To what extent do teachers agree on AI's perceived effectiveness, drawbacks, and their willingness to integrate it into pronunciation teaching?

b) How does the use of AI in pronunciation teaching vary based on factors such as age, years of experience, education, the level of education they teach, their school's sector, and prior training?

c) How do teachers' beliefs about using AI for pronunciation teaching differ based on these factors, including their actual use of AI?

The first research question will be answered by employing descriptive statistics from the survey regarding the beliefs of teachers about AI for pronunciation teaching. The second research question will be answered by identifying differences between different groups of teachers in terms of AI use. To answer the third research question, potential differences between various groups of teachers in terms of their beliefs on AI will be examined. The findings can inform teacher training programs, policy decisions, and future research on the effective integration of AI in pronunciation instruction. The following hypotheses are formed based on the research questions:

H1: Based on existing literature, teachers are likely to agree significantly on the effectiveness of AI and show a willingness to integrate it into their teaching. However, there may be lower levels of agreement regarding the perceived drawbacks of AI for pronunciation instruction.

H2: Given that AI is a novel concept for many, and considering prior findings, AI use may not significantly vary across different teacher demographics. However, since training influences teachers' intentions to use AI, it is anticipated that teachers with more extensive training will incorporate AI more frequently than those with less training.

H3: While it is challenging to predict the extent of variation in beliefs across these factors due to the lack of prior research, the strong connection between teachers' beliefs and practices, coupled with the significant role of training in AI adoption, suggests that differences in beliefs may emerge. Specifically, teachers with greater AI use and training are expected to have more favorable views on its effectiveness and greater openness to integration, while also being less likely to emphasize its drawbacks.

## 2. Methodology

### 2.1 Participants

The study involved 117 in-service EFL teachers ($n$females = 106) working in Cyprus. Participants' ages ranged from 22 to 64 years ($M = 43.62$, $SD = 9.96$), with an average of 18.29 years ($SD = 9.56$) of EFL teaching experience. The majority of participants were native Greek speakers ($n = 103$). Regarding educational qualifications, 32 teachers held a Bachelor's (BA) degree, 72 held a Master's (MA) degree, and 13 had a Doctoral (PhD) degree. In terms of workplace distribution, 15 teachers worked in primary education, 78 in secondary education, and 24 in adult education. Sixty were employed in private schools, while the remaining 57 worked in public schools. All teachers stated an average use of AI for pronunciation teaching of 2.64 ($SD = 1.75$) and reception of training of 1.85 ($SD = 1.39$); these were measured on a Likert-point scale from 1–7. For the investigation of how various teacher demographic and professional factors affect the use of AI, teachers were divided into different groups based on their age (younger, older), experience (low, high), education (BA, MA, PhD), level of the school they work (primary, secondary, adult education), sector of the school they work (public, private) and training they received (low, moderate, high). A k-means clustering technique was applied to segment participants based on their age and experience (measured in years). To implement the clustering, the number of clusters was set to 2, with the goal of identifying two distinct age and experience groups within the dataset. The k-means algorithm grouped the participants into two clusters, and each participant was assigned to one of these clusters based on their age and experience. The average age and experience within each cluster were then calculated using the tapply() function, and the groups with the lower mean age and experience were labeled as the "Younger" and "High" groups, while the other clusters were designated as the "Older" and "Low" groups.

### 2.2 Tool

An online survey was used for data collection. The survey consisted of two main parts. In the first part, participants provided details such as age, gender, native language, level of education they work, sector in which their school belongs, years of experience, the extent to which they use AI in their EFL classrooms (1 = never, 7 = always), and the extent to which they received training on the use of AI for pronunciation teaching (1 = not at all, 7 = very extensive). In the second part, participants responded to statements related to beliefs about AI in pronunciation teaching. More specifically, these statements assessed agreement (1 = strongly disagree, 7 = strongly agree) with three statements on perceived effectiveness, three on perceived drawbacks, and three on openness toward AI integration in pronunciation teaching. The exact statements are presented in Table 1. To investigate the reliability of the questionnaire, Cronbach's alpha was calculated, yielding a value of 0.76, indicating acceptable internal consistency.

Table 1: Statements reflecting teachers' beliefs about AI in pronunciation teaching.

| Beliefs | Statements |
| --- | --- |
| Perceived effectiveness | 1. I believe AI is an effective tool for pronunciation teaching.<br>2. I believe AI can save teachers time in providing pronunciation feedback.<br>3. I believe AI-based pronunciation exercises are engaging for students. |
| Perceived drawbacks | 4. I am concerned that AI may replace the human element in pronunciation teaching.<br>5. I have concerns about the accuracy of AI for pronunciation analysis.<br>6. I have concerns about the collection and use of student data by AI systems. |
| AI integration | 7. I am open to incorporating AI tools into my regular pronunciation instruction.<br>8. I am interested in learning more about the potential of AI in pronunciation teaching.<br>9. I would be willing to participate in professional training workshops on using AI for pronunciation instruction. |

## 2.3 Procedure

The survey was administered online via Google Forms and distributed through social media posts and direct email invitations to EFL teachers across Cyprus. The survey's language was English. There were no time constraints for responses. Before completing the survey, participants were informed about the study's purpose, participation criteria, and ethical considerations. Informed consent was obtained. The estimated completion time for the survey was approximately 10 minutes.

## 3. Results

### 3.1 Beliefs on the use of AI for pronunciation teaching

At first glance, as seen in Figure 1, the findings show there is a higher degree of agreement about the willingness of teachers to integrate AI in pronunciation teaching followed by aspects of perceived effectiveness and finally concerns about AI.

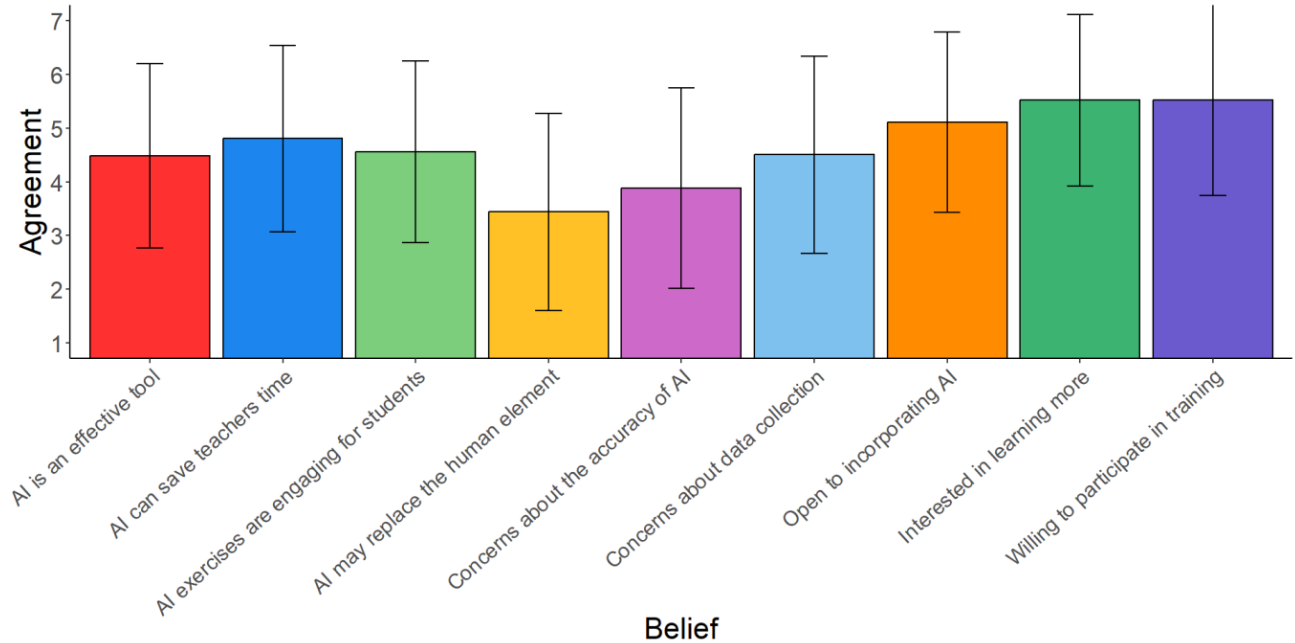

Figure 1: Teachers' beliefs about the AI for pronunciation teaching

To investigate the differences across the degree of agreement for the beliefs, an ordinal regression model was employed using the MASS package (Ripley & Venables, 2025) in R (R Core Team, 2025). Agreement was modeled as the dependent variable (Likert-point scale from 1 to 7), while Belief was modeled as the independent variable. The results exhibited significant differences between the beliefs "AI may replace the human element", "Concerns about the accuracy of AI", all three beliefs regarding teachers' openness to incorporate AI, and the baseline belief (i.e., "AI is an effective tool" ($\beta$s = –1.02–1.22, $SE$s = 0.23–0.24, $t$s = –4.40–5.16, $p$s = 0.01 – < 0.001). Subsequent posthoc tests with the Tukey correction revealed significant differences notably between the majority of the beliefs regarding perceived effectiveness (i.e., Beliefs 1–3), openness to incorporate AI (i.e., Beliefs 7–9), and beliefs regarding perceived drawbacks (i.e., Beliefs 4–6), with a higher degree of agreement for the former two clusters. In addition, significant differences occurred between the majority of the beliefs regarding perceived effectiveness and the majority of the beliefs about openness to incorporate AI, with the latter concentrating higher degree of agreement. The detailed results of the posthoc test are presented in Table 2.

Table 2: Results of the pairwise comparisons for differences between the degree of agreement among various beliefs. Only pairs with significant differences are presented.

| Belief pairs* | Estimate | Std Error | *t*-value | *p*-value |
|---|---|---|---|---|
| 1 – 4 | 1.02 | 0.23 | 4.40 | <0.01 |
| 1 – 6 | –0.04 | 0.23 | –0.19 | 1.00 |
| 1 – 8 | –1.10 | 0.23 | –4.73 | <0.01 |
| 1 – 9 | –1.22 | 0.24 | –5.16 | <.0001 |
| 2 – 4 | 1.34 | 0.23 | 5.72 | <.0001 |
| 2 – 5 | 0.88 | 0.23 | 3.80 | <0.01 |
| 2 – 8 | –0.78 | 0.23 | –3.36 | 0.02 |
| 2 – 9 | –0.90 | 0.24 | –3.81 | <0.01 |
| 3 – 4 | 1.09 | 0.23 | 4.72 | <0.01 |
| 3 – 8 | –1.03 | 0.23 | –4.44 | <0.01 |
| 3 – 9 | –1.15 | 0.24 | –4.87 | <.0001 |
| 4 – 6 | –1.06 | 0.23 | –4.53 | <0.01 |
| 4 – 7 | –1.65 | 0.24 | –6.97 | <.0001 |
| 4 – 8 | –2.12 | 0.24 | –8.79 | <.0001 |
| 4 – 9 | –2.24 | 0.25 | –9.12 | <.0001 |
| 5 – 7 | –1.19 | 0.23 | –5.10 | <.0001 |
| 5 – 8 | –1.66 | 0.24 | –6.98 | <.0001 |
| 5 – 9 | –1.78 | 0.24 | –7.36 | <.0001 |
| 6 – 8 | –1.06 | 0.24 | –4.51 | <0.01 |
| 6 – 9 | –1.18 | 0.24 | –4.93 | <.0001 |

*1 = AI is an effective tool; 2 = AI can save teachers time; 3 = AI exercises are engaging for students; 4 = AI may replace the human element; 5 = Concerns about the accuracy of AI, 6 = Concerns about data collection; 7 = Open to incorporating AI, 8 = Interested in learning more; 9 = Willing to participate in training.

## 3.2 Use of AI and the effect of different factors

The descriptive statistics revealed that younger teachers and those with less teaching experience tend to use AI more frequently for pronunciation teaching. Additionally, teachers holding an MA degree were found to use AI more frequently than those with BA and PhD degrees. Among the different educational sectors, adult education teachers demonstrated the highest use of AI, followed by secondary and primary education teachers. AI use appeared similar between teachers working in public and private schools. Moreover, teachers receiving more extensive training used AI more frequently than those with less training. Figure 2 illustrates the extent of AI use for pronunciation teaching across various teacher demographic and professional factors.

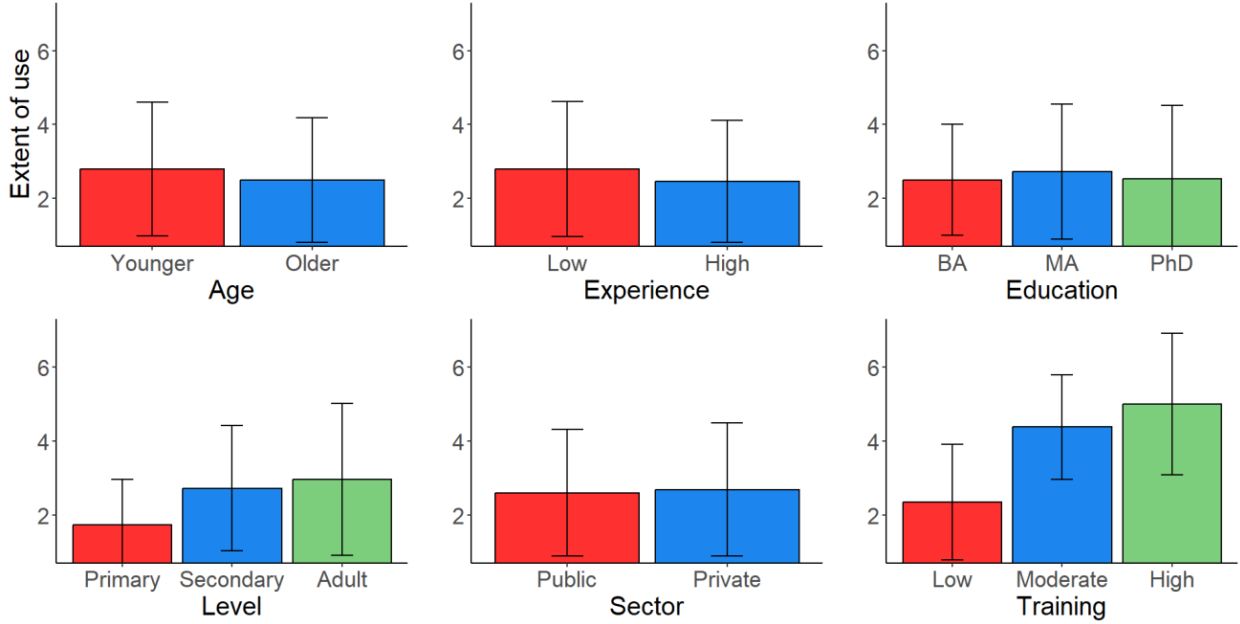

Figure 2: Extent of AI use for pronunciation teaching (Likert scale 1–7; 1 = never, 7 = always) across various teacher demographic and professional factors.

The statistical analysis was conducted using an ordinal logistic regression model in R. Use, which reflected the extent to which teachers use AI on a Likert-point scale from 1 to 7) was modeled as the dependent variable, whereas other variables including Age, Experience, Education, Level, Sector, and Training, were modeled as independent. The analysis indicated no significant differences between the AgeOlder, ExperienceHigh, and the baseline variables, suggesting that age and teaching experience did not influence AI use. Moreover, no significant differences were found between EducationMA, EducationPhD, and the baseline variable, indicating that educational background did not have an impact on AI use. However, significant differences were observed for LevelSecondary, LevelAdult, and the baseline, highlighting that teachers working with adolescents and adults used AI more frequently than those teaching children. No significant differences were found between teachers in private schools (SectorPrivate) and the baseline. Finally, significant differences were noted between the TrainingModerate, TrainingHigh categories, and the baseline variable, indicating that more extensive training was associated with more frequent use of AI for pronunciation teaching. The results of the model are presented in Table 3.

Table 3: Results of the ordinal regression logistics model regarding differences in AI use based across various teacher demographic and professional factors. The baseline variables are AgeYounger, ExperienceLow, EducationBA, LevelPrimary, SectorPublic, and TrainingLow.

|  | **Estimate** | **Std Error** | ***t*–value** | ***p*–value** |
|---|---|---|---|---|
| AgeOlder | 0.20 | 0.58 | 0.35 | 0.73 |
| ExperienceHigh | –0.57 | 0.55 | –1.03 | 0.30 |
| EducationMA | 0.21 | 0.41 | 0.51 | 0.61 |
| EducationPhD | –0.43 | 0.67 | –0.64 | 0.52 |
| LevelSecondary | 1.30 | 0.60 | 2.15 | 0.03 |
| LevelAdult | 1.45 | 0.70 | 2.08 | 0.04 |
| SectorPrivate | 0.17 | 0.42 | 0.41 | 0.69 |
| TrainingModerate | 1.80 | 0.64 | 2.82 | < 0.001 |
| TrainingHigh | 2.93 | 0.80 | 3.66 | < 0.001 |

### 3.3 Beliefs about AI and the effect of different factors

The descriptive statistics revealed that teachers who use AI more frequently in pronunciation teaching tend to have a stronger agreement regarding the perceived effectiveness of AI and a greater willingness to integrate it into their practices. However, a similar trend is not observed in relation to perceived drawbacks (see Figure 3). Age, experience, and sector do not appear to significantly influence teachers' beliefs, as shown in Figures 4, 5, and 8. Regarding educational background, teachers with a BA tend to have lower agreement on the perceived effectiveness of AI and higher agreement on its perceived drawbacks compared to those with higher degrees (see Figure 6). Additionally, primary education teachers show slightly lower agreement on the effectiveness of AI compared to their secondary and adult education counterparts (see Figure 7). Finally, teachers with higher levels of AI training for pronunciation teaching tend to express fewer concerns about AI than those with lower levels of training (see Figure 9).

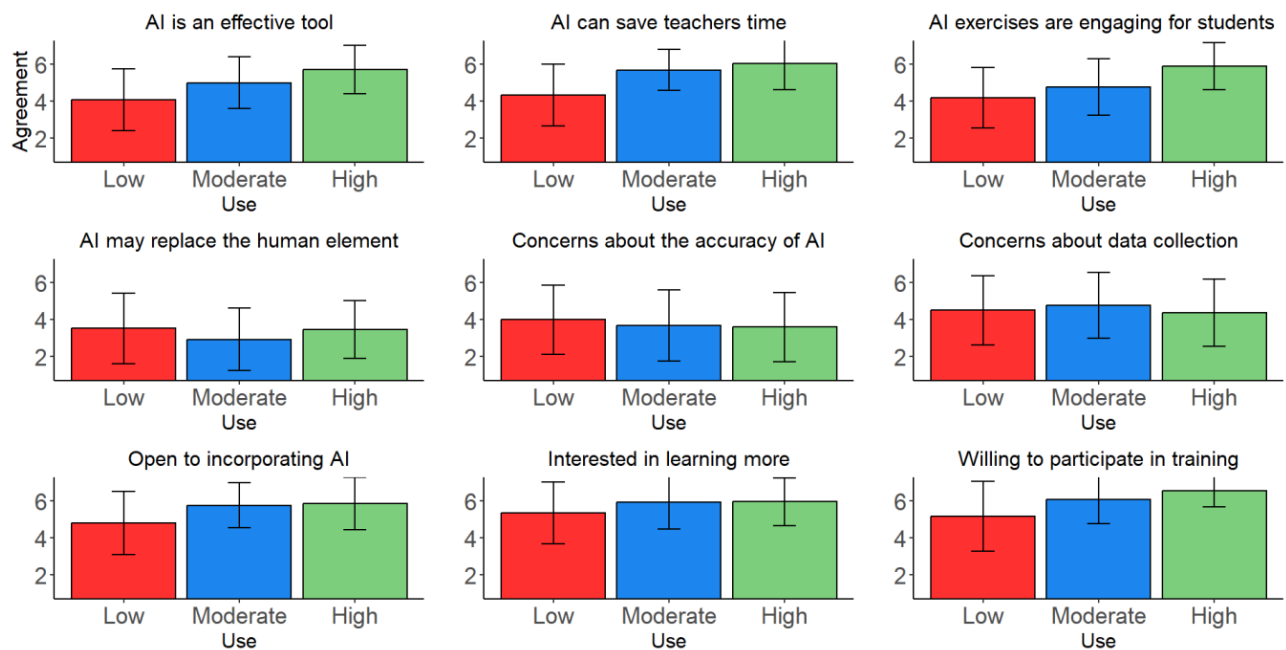

Figure 3: Teachers' degree of agreement on various beliefs about AI in pronunciation teaching by use level.

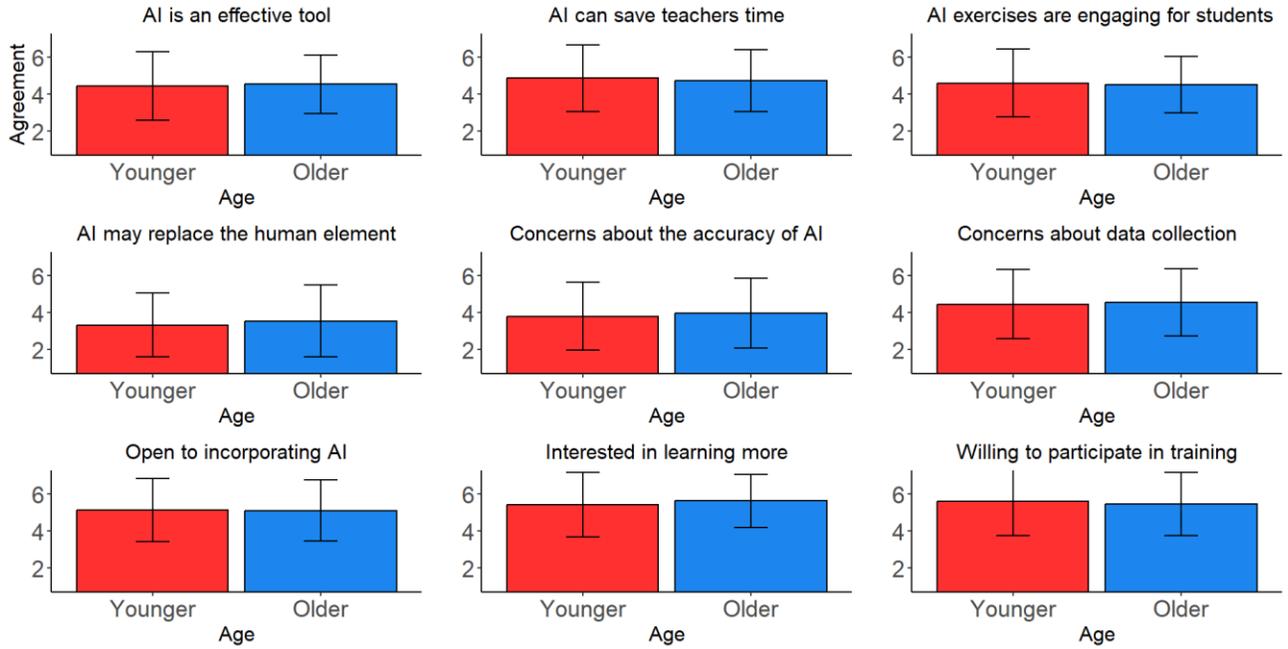

Figure 4: Teachers' degree of agreement on various beliefs about AI in pronunciation teaching by age.

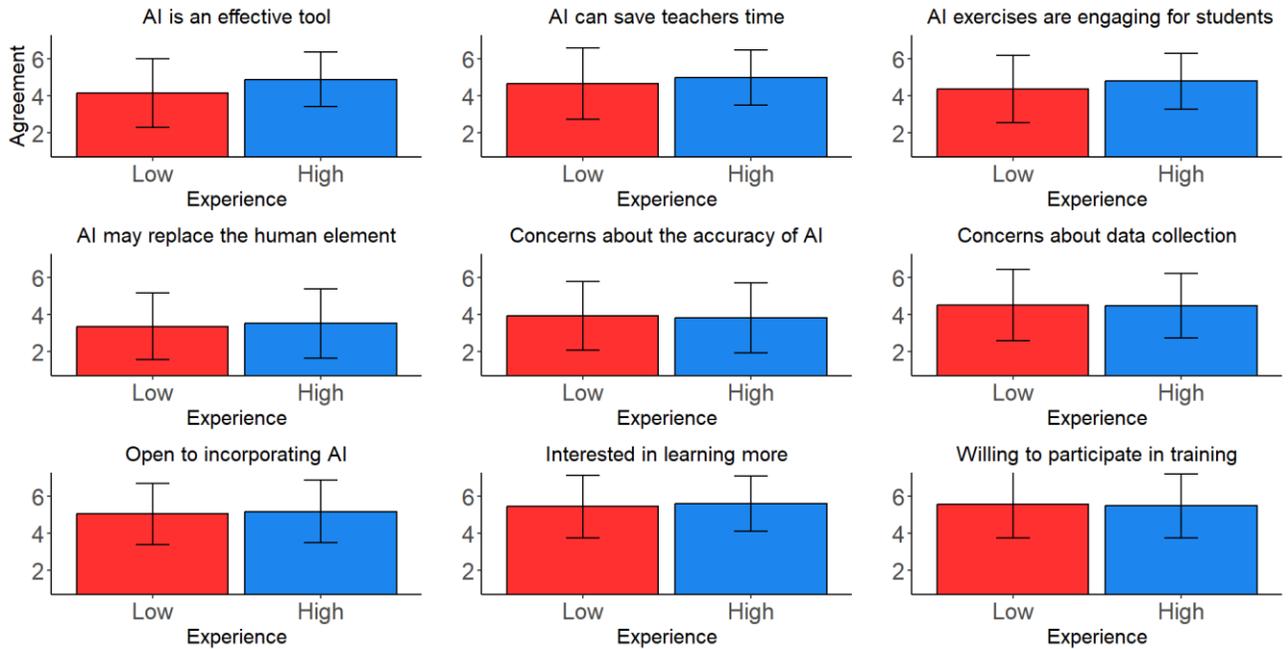

Figure 5: Teachers' degree of agreement on various beliefs about AI in pronunciation teaching by experience.

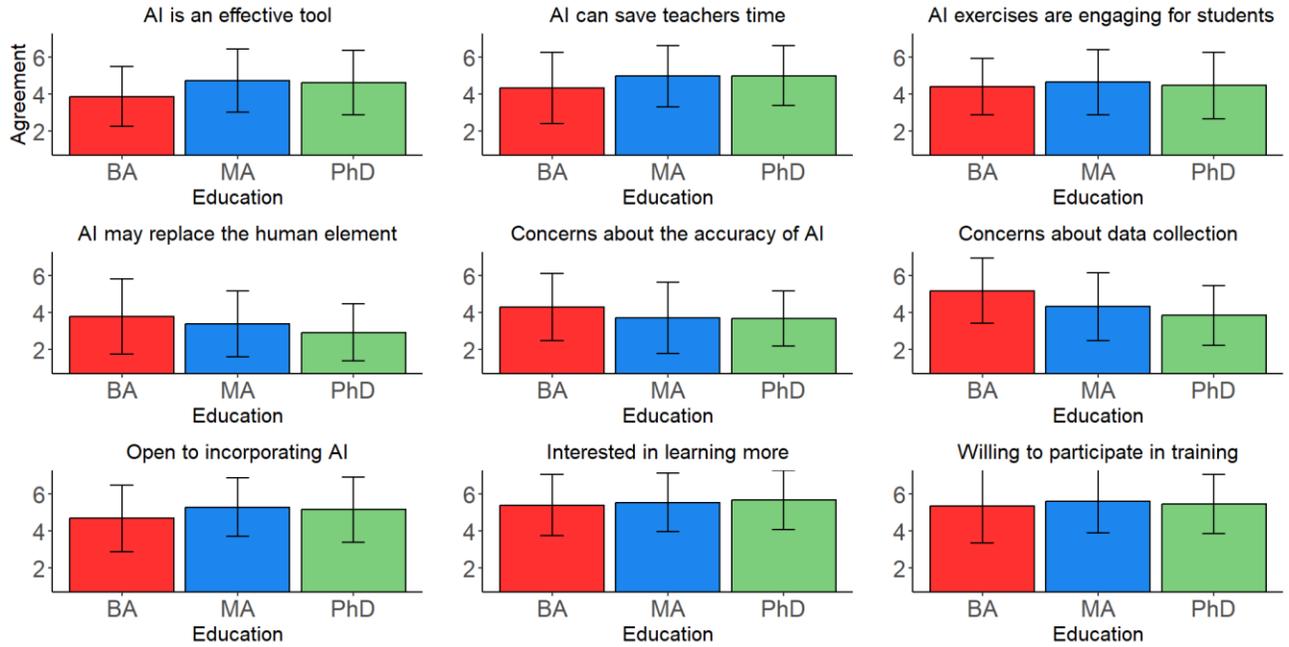

Figure 6: Teachers' degree of agreement on various beliefs about AI in pronunciation teaching by their educational backgrounds.

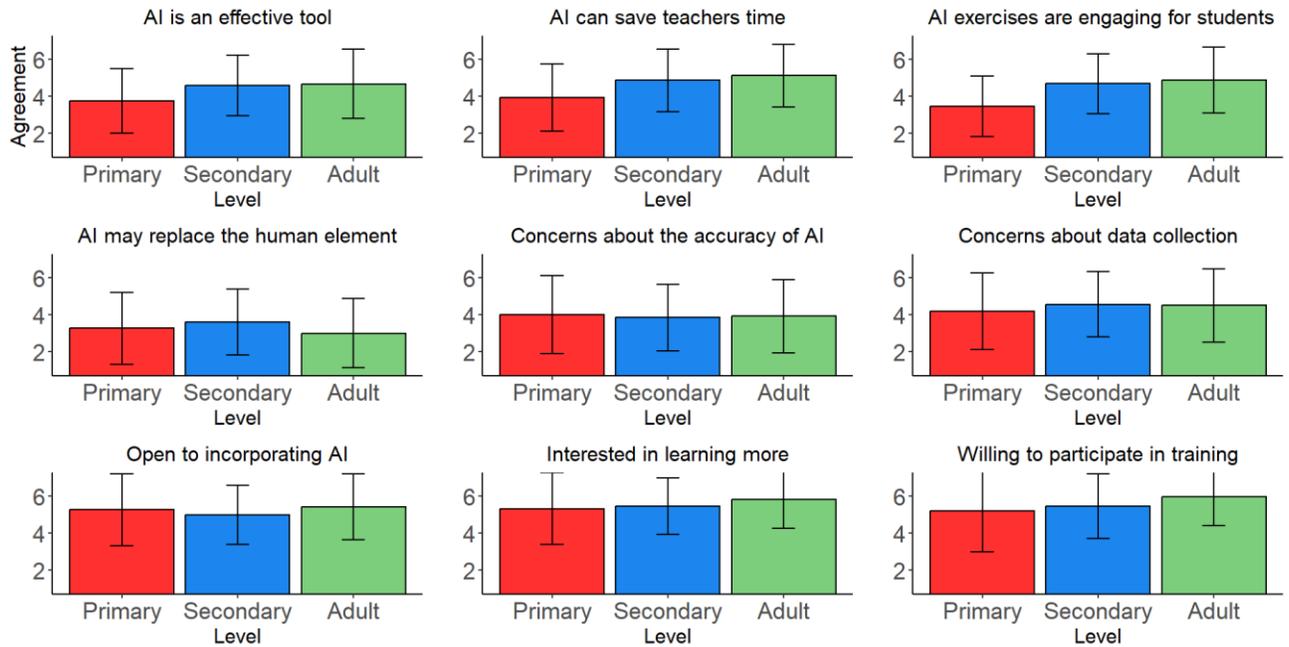

Figure 7: Teachers' degree of agreement on various beliefs about AI in pronunciation teaching by the educational level they teach.

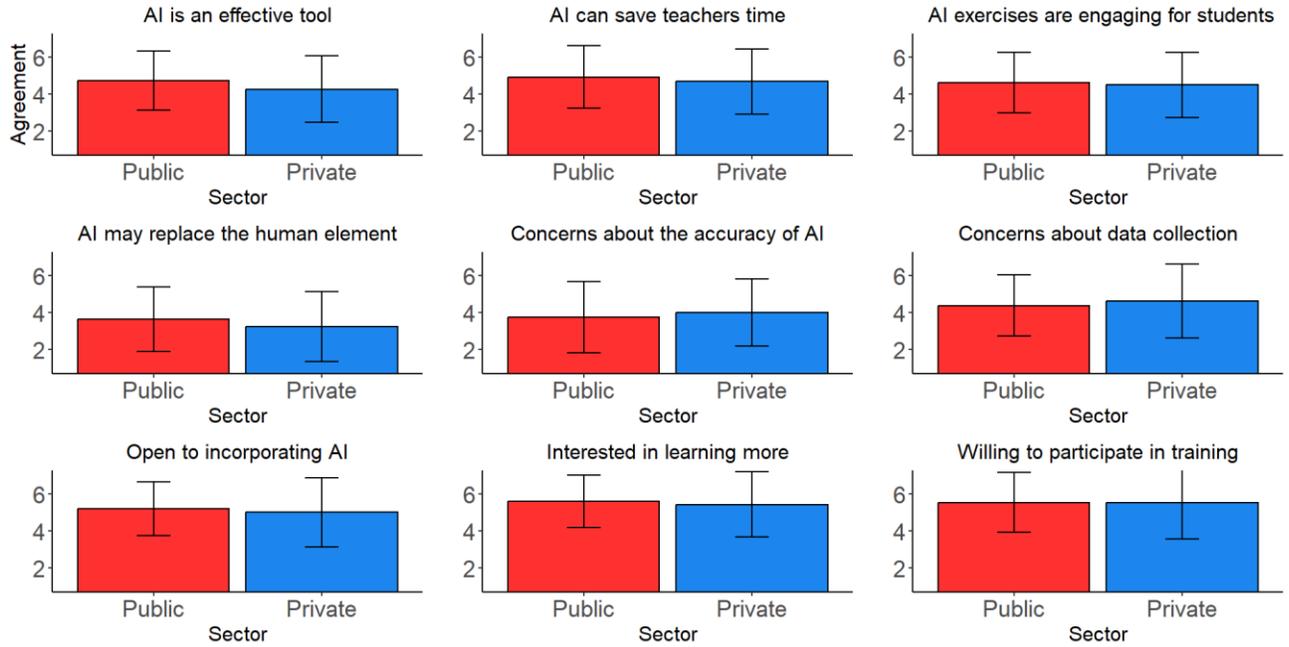

Figure 8: Teachers' degree of agreement on various beliefs about AI in pronunciation teaching by their schools' sectors.

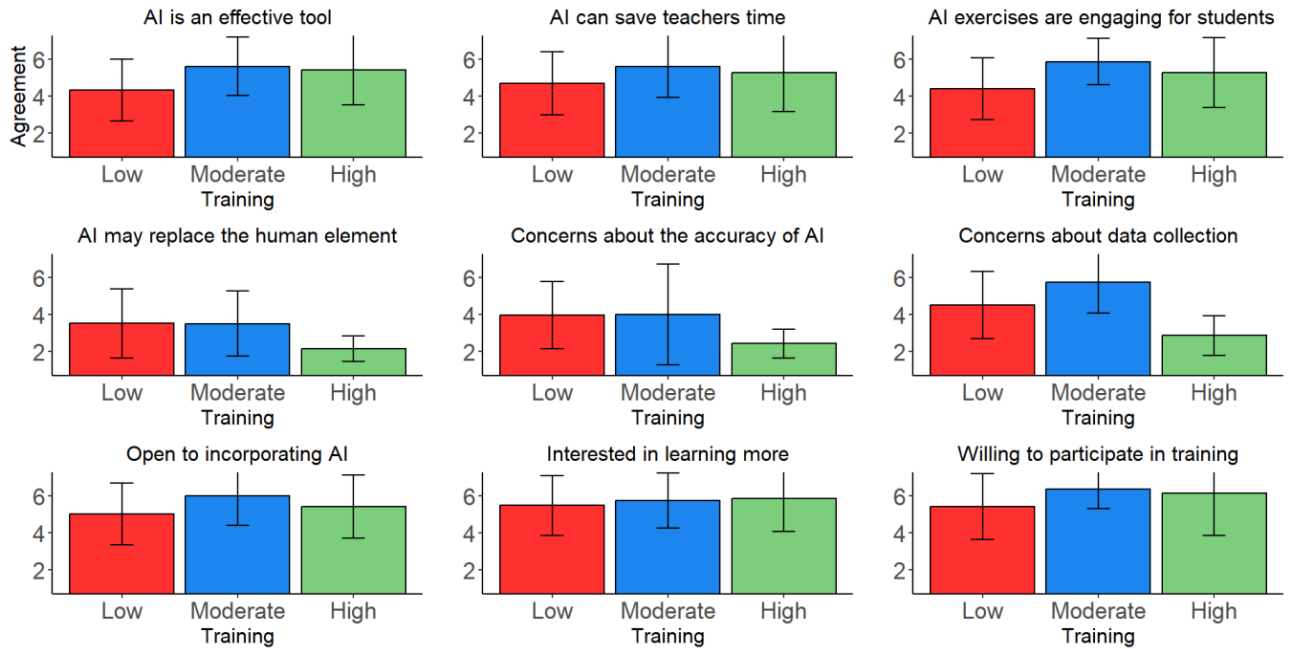

Figure 9: Teachers' degree of agreement on various beliefs about AI in pronunciation teaching by the extent to which they received relevant training.

A series of ordinal logistic regression models was conducted in R to examine potential differences in teachers' beliefs about AI in pronunciation teaching, based on various factors. The results revealed significant differences between UseHigh and the baseline variable in terms of perceived effectiveness and willingness to incorporate AI; thus, teachers with higher AI use held more positive beliefs regarding the effectiveness of AI and were more willing to use it in their teaching. However, no significant differences were observed regarding perceived drawbacks. Regarding age, significant differences were found only for the belief that "AI exercises are engaging for students", with older teachers showing stronger agreement. For experience, significant differences emerged for the beliefs that "AI is an effective tool" and "AI exercises are engaging for students", with highly experienced teachers showing stronger agreement compared to their less experienced counterparts. In terms of education, EducationMA significantly differed from the baseline category regarding the belief in the effectiveness of AI, with MA holders showing stronger agreement than BA holders. Moreover, differences were found between MA, PhD holders, and the baseline variable regarding concerns about data collection and use, with the former holders expressing less concern. No significant differences were observed based on the educational level they teach or the sector of their schools. Finally, significant differences were found between TrainingHigh and the baseline group on the beliefs "AI may replace the human element" and "Concerns about data collection"; teachers with more extensive training were less likely to agree with these beliefs. The results of the models are shown in Table 4.

Table 4: Results of the ordinal regression logistics models regarding teachers' degree of agreement on beliefs about AI in pronunciation teaching across various factors. The baseline variables are UseLow, AgeOlder, ExperienceLow, EducationBA, LevelPrimary, SectorPublic, and TrainingLow.

| Belief | Variable | Estimate | Std Error | $t$–value | $p$–value |
|---|---|---|---|---|---|
| **AI is an effective tool** | UseModerate | 1.03 | 0.57 | 1.79 | 0.07 |
| | UseHigh | 1.86 | 0.51 | 3.68 | <0.001 |
| | AgeOlder | –1.11 | 0.60 | –1.83 | 0.07 |
| | ExperienceHigh | 1.64 | 0.55 | 2.97 | <0.001 |
| | EducationMA | 0.82 | 0.41 | 1.97 | 0.05 |
| | EducationPhD | 0.50 | 0.67 | 0.74 | 0.46 |
| | LevelSecondary | 0.26 | 0.56 | 0.47 | 0.64 |
| | LevelAdult | 0.74 | 0.68 | 1.09 | 0.27 |
| | SectorPrivate | –0.72 | 0.43 | –1.65 | 0.10 |
| | TrainingModerate | 0.32 | 0.78 | 0.42 | 0.68 |
| | TrainingHigh | 0.18 | 0.82 | 0.23 | 0.82 |
| **AI can save teachers time** | UseModerate | 1.39 | 0.59 | 2.36 | 0.02 |
| | UseHigh | 2.32 | 0.56 | 4.16 | <0.001 |
| | AgeOlder | –0.71 | 0.64 | –1.11 | 0.27 |
| | ExperienceHigh | 0.73 | 0.58 | 1.26 | 0.21 |
| | EducationMA | 0.46 | 0.42 | 1.09 | 0.28 |
| | EducationPhD | 0.34 | 0.67 | 0.51 | 0.61 |
| | LevelSecondary | 0.66 | 0.56 | 1.19 | 0.24 |
| | LevelAdult | 1.03 | 0.68 | 1.52 | 0.13 |
| | SectorPrivate | –0.58 | 0.43 | –1.35 | 0.18 |
| | TrainingModerate | –0.22 | 0.78 | –0.29 | 0.78 |
| | TrainingHigh | –0.75 | 0.78 | –0.96 | 0.34 |
| **AI exercises are engaging for students** | UseModerate | 0.26 | 0.56 | 0.46 | 0.64 |
| | UseHigh | 1.96 | 0.51 | 3.82 | <0.001 |
| | AgeOlder | –1.16 | 0.59 | –1.97 | 0.05 |
| | ExperienceHigh | 1.23 | 0.55 | 2.25 | 0.02 |
| | EducationMA | 0.01 | 0.40 | 0.02 | 0.98 |
| | EducationPhD | –0.47 | 0.66 | –0.70 | 0.48 |
| | LevelSecondary | 0.77 | 0.54 | 1.41 | 0.16 |
| | LevelAdult | 1.20 | 0.65 | 1.84 | 0.07 |
| | SectorPrivate | –0.31 | 0.44 | –0.70 | 0.48 |
| | TrainingModerate | 0.67 | 0.72 | 0.93 | 0.35 |
| | TrainingHigh | –0.23 | 0.83 | –0.28 | 0.78 |
| **AI may replace the human element** | UseModerate | –0.51 | 0.58 | –0.87 | 0.38 |
| | UseHigh | 0.67 | 0.50 | 1.34 | 0.18 |
| | AgeOlder | –0.34 | 0.58 | –0.59 | 0.56 |
| | ExperienceHigh | <0.001 | 0.53 | 0.01 | 1.00 |
| | EducationMA | –0.41 | 0.42 | –0.97 | 0.33 |
| | EducationPhD | –0.42 | 0.61 | –0.69 | 0.49 |
| | LevelSecondary | 0.30 | 0.56 | 0.54 | 0.59 |
| | LevelAdult | –0.22 | 0.65 | –0.34 | 0.73 |
| | SectorPrivate | –0.61 | 0.43 | –1.42 | 0.15 |
| | TrainingModerate | 0.21 | 0.70 | 0.30 | 0.77 |
| | TrainingHigh | –1.97 | 0.77 | –2.55 | 0.01 |
| **Concerns about the accuracy of AI** | UseModerate | –0.28 | 0.60 | –0.47 | 0.64 |
| | UseHigh | 0.09 | 0.48 | 0.18 | 0.86 |
| | AgeOlder | 0.56 | 0.58 | 0.97 | 0.33 |
| | ExperienceHigh | –0.42 | 0.53 | –0.79 | 0.43 |
| | EducationMA | –0.51 | 0.40 | –1.26 | 0.21 |
| | EducationPhD | –0.34 | 0.60 | –0.56 | 0.57 |
| | LevelSecondary | –0.05 | 0.58 | –0.09 | 0.93 |
| | LevelAdult | –0.18 | 0.67 | –0.27 | 0.79 |
| | SectorPrivate | 0.34 | 0.43 | 0.80 | 0.42 |
| | TrainingModerate | 0.53 | 0.92 | 0.57 | 0.57 |
| | TrainingHigh | –1.27 | 0.72 | –1.76 | 0.08 |
| **Concerns about data collection** | UseModerate | –0.36 | 0.58 | –0.63 | 0.53 |
| | UseHigh | 0.18 | 0.50 | 0.37 | 0.71 |
| | AgeOlder | 0.31 | 0.55 | 0.57 | 0.57 |
| | ExperienceHigh | –0.31 | 0.50 | –0.61 | 0.54 |
| | EducationMA | –1.00 | 0.42 | –2.37 | 0.02 |
| | EducationPhD | –1.22 | 0.63 | –1.94 | 0.05 |
| | LevelSecondary | 0.41 | 0.58 | 0.71 | 0.48 |
| | LevelAdult | 0.19 | 0.68 | 0.28 | 0.78 |
| | SectorPrivate | 0.36 | 0.41 | 0.88 | 0.38 |
| | TrainingModerate | 1.66 | 0.75 | 2.22 | 0.03 |
|  | TrainingHigh | –1.71 | 0.76 | –2.25 | 0.02 |
| **Open to incorporating AI** | UseModerate | 1.03 | 0.59 | 1.76 | 0.08 |
|  | UseHigh | 1.58 | 0.58 | 2.75 | 0.01 |
|  | AgeOlder | –0.27 | 0.60 | –0.46 | 0.64 |
|  | ExperienceHigh | 0.64 | 0.56 | 1.14 | 0.25 |
|  | EducationMA | 0.28 | 0.41 | 0.68 | 0.50 |
|  | EducationPhD | 0.29 | 0.70 | 0.42 | 0.68 |
|  | LevelSecondary | –1.07 | 0.58 | –1.84 | 0.07 |
|  | LevelAdult | –0.26 | 0.70 | –0.38 | 0.71 |
|  | SectorPrivate | –0.50 | 0.44 | –1.14 | 0.26 |
|  | TrainingModerate | 0.46 | 0.82 | 0.56 | 0.58 |
|  | TrainingHigh | –0.49 | 0.83 | –0.59 | 0.56 |
| **Interested in learning more** | UseModerate | 0.95 | 0.64 | 1.49 | 0.14 |
|  | UseHigh | 0.74 | 0.54 | 1.37 | 0.17 |
|  | AgeOlder | 0.16 | 0.59 | 0.27 | 0.79 |
|  | ExperienceHigh | 0.13 | 0.55 | 0.24 | 0.81 |
|  | EducationMA | 0.05 | 0.41 | 0.11 | 0.91 |
|  | EducationPhD | <0.001 | 0.67 | –0.01 | 0.99 |
|  | LevelSecondary | –0.30 | 0.56 | –0.53 | 0.60 |
|  | LevelAdult | 0.46 | 0.68 | 0.68 | 0.50 |
|  | SectorPrivate | –0.15 | 0.43 | –0.36 | 0.72 |
|  | TrainingModerate | –0.31 | 0.77 | –0.40 | 0.69 |
|  | TrainingHigh | 0.10 | 0.86 | 0.12 | 0.90 |
| **Willing to participate in training** | UseModerate | 0.83 | 0.64 | 1.31 | 0.19 |
|  | UseHigh | 1.46 | 0.59 | 2.48 | 0.01 |
|  | AgeOlder | –0.26 | 0.60 | –0.43 | 0.67 |
|  | ExperienceHigh | 0.22 | 0.55 | 0.41 | 0.68 |
|  | EducationMA | –0.01 | 0.44 | –0.01 | 0.99 |
|  | EducationPhD | –0.69 | 0.67 | –1.04 | 0.30 |
|  | LevelSecondary | –0.20 | 0.59 | –0.33 | 0.74 |
|  | LevelAdult | 0.73 | 0.71 | 1.03 | 0.30 |
|  | SectorPrivate | –0.15 | 0.44 | –0.35 | 0.73 |
|  | TrainingModerate | 0.15 | 0.84 | 0.18 | 0.86 |
|  | TrainingHigh | 1.13 | 1.25 | 0.91 | 0.36 |

## 4. Discussion

This study investigated the use of AI by EFL teachers in Cyprus for pronunciation teaching and their beliefs about AI. More specifically, it examined how AI use varies based on different teacher demographic and professional factors, and how the beliefs of teachers vary based on these factors including AI use. This is the first study focusing on the use of AI and the beliefs of teachers in the underresearched area of pronunciation instruction.

The first aim of this study was to examine the extent to which teachers agree on the perceived effectiveness and drawbacks of AI, as well as their willingness to integrate it into pronunciation teaching. The results confirm H1, suggesting that teachers showed greater agreement on the effectiveness of AI and their willingness to use it for pronunciation instruction compared to its perceived drawbacks. This aligns with findings from numerous studies that highlight positive perceptions of AI use among EFL teachers (Dahia, 2024; Sumakul, 2022). The high willingness to adopt AI is also consistent with prior research (e.g., An et al., 2023; Sütçü & Sütçü, 2023). This may be attributed to teachers' beliefs that AI can be an effective tool for pronunciation instruction, save time for teachers, and provide engaging activities for students. Teachers appear to be less concerned about the potential replacement of the human element by AI, likely because such a scenario currently seems unrealistic. Nevertheless, previous literature shows that this concern was more pronounced among teachers before the commercial introduction of AI (Haseski, 2019). In fact, research suggests a transformation of the profession rather than a complete replacement of teachers (Popenici & Kerr, 2017), a viewpoint that is acknowledged by EFL teachers. The greatest concern among teachers was the collection and use of students' personal information, a valid concern given the potential risks to privacy for both students and teachers (Nguyen, 2023); similarly, Aljemely (2024) argued that teachers believe that AI use may compromise their privacy and give rise to ethical issues. However, these concerns can be mitigated through proper training that educates teachers on how data is collected, stored, and processed within AI-powered systems (Pardo & Siemens, 2014).

The second research question examined differences in AI use for pronunciation teaching among teachers based on age, years of experience, education, the level of education they teach, their school's sector, and prior training. As predicted by H2, neither age nor experience significantly influenced AI use, aligning with the findings of Cabero-Almenara et al. (2024) and Alghamdi (2023). This similarity was expected, given the strong correlation between age and experience ($r = 0.82$, $p < 0.001$). The widespread availability and accessibility of AI-powered pronunciation tools likely explain this lack of variation, as teachers of all ages may have similar exposure to these technologies. Additionally, training and professional development opportunities may play a more decisive role in AI adoption than age alone, particularly since younger and older teachers received comparable amounts of training ($t = -0.78$, df $= 115$, $p = 0.43$). Similarly, teachers' educational backgrounds did not appear to influence AI use. Likewise, whether a teacher worked in the public or private sector had no significant effect, possibly due to comparable levels of institutional support and resources – key factors in AI adoption (Molefi et al., 2024). However, differences emerged based on the level of education teachers worked in. Secondary and adult education teachers reported more frequent AI use than their primary school counterparts. This finding may be linked to the lesser emphasis placed on pronunciation instruction in primary education compared to higher educational levels (Szyszka, 2016). AI adoption may be more prevalent in adult education due to the greater demand for pronunciation training for professional and academic purposes. Finally, as projected by H2, teachers with higher levels of training used AI more frequently than those with less training. The acquisition of digital skills and technological proficiency appears to facilitate AI integration, equipping teachers with the necessary knowledge to effectively utilize these tools and overcome potential challenges (Aljemery, 2024; Molefi et al., 2024).

The third research question examined how teachers' beliefs about using AI for pronunciation instruction varied based on the above factors, including the extent of AI use. As hypothesized in H3, increased AI use was linked to a stronger agreement on the perceived effectiveness of AI and a greater willingness among teachers to incorporate it into their practices. This finding aligns with previous studies that reveal the connection between the perceived usefulness of AI and more favorable attitudes towards its use in education (An et al., 2023; Zhang et al., 2023). Interestingly, older and more experienced teachers expressed stronger agreement regarding the effectiveness of AI compared to their younger and less experienced counterparts. It is possible that these teachers, having observed the limitations of traditional methods over the years, are more open to adopting innovative techniques that could enhance pronunciation instruction. Furthermore, teachers with higher academic qualifications (i.e., MA or PhD) were more likely to perceive AI as effective and raised fewer concerns about data collection and privacy issues than those with only a BA. These results agree with Arowosegbe et al. (2024), who found that postgraduate students held more positive perceptions of AI compared to their undergraduate peers. This may be attributed to the advanced digital and research literacy typically fostered by postgraduate education. Finally, teachers who had received more extensive training in AI showed fewer concerns compared to those with less training, supporting H3. It is understandable that teachers who lack knowledge about how AI operates may be wary of issues related to data privacy. However, as demonstrated in this study, training can foster trust and understanding (Nazaretsky et al., 2022), helping educators to better navigate concerns about AI implementation (Viberg et al., 2024).

The findings of this study have significant pedagogical implications for enhancing the acceptability and effectiveness of AI in pronunciation teaching, ultimately leading to improved learning outcomes. One of the key takeaways is the critical role of training and professional development in fostering the adoption of AI tools for pronunciation instruction. It is essential for relevant ministries and educational institutions to prioritize comprehensive training programs that address both the technical aspects of AI tools and ethical considerations, such as data privacy. This will empower educators to feel more confident and competent in incorporating AI into their teaching practices. Training programs should also be tailored to meet the diverse needs of teachers, accounting for varying levels of academic qualifications and professional experience. In particular, for the latter, strategies should be developed to encourage the use of AI for pronunciation teaching in primary schools. This can be achieved by raising awareness about the importance of pronunciation in early education and incorporating AI tools designed specifically for younger learners. Furthermore, it is crucial for schools in both the public and private sectors to ensure equitable access to resources and technologies that facilitate the integration of AI in the classroom. This may involve providing financial support for AI tools, offering technical assistance, and fostering collaborative initiatives between institutions to share best practices. Ultimately, the link between increased AI use and a stronger belief in its effectiveness underscores the importance of cultivating favorable perceptions of AI among educators. Educational programs and initiatives should emphasize the tangible benefits of AI in enhancing pronunciation instruction, thereby encouraging its wider adoption and integration into teaching practices.

## 5. Conclusion

The present study provides significant insights into the use of AI in pronunciation instruction and the beliefs of EFL teachers working in Cypriot schools. The findings contribute to the ongoing discourse on the role of AI in language education, particularly in pronunciation instruction, and can serve as a valuable resource for informing teacher training programs, shaping educational policies, and guiding future research in this area. The results demonstrate the potential of AI to enhance pronunciation teaching but also indicate the importance of understanding teachers' perceptions to fully harness its benefits. However, this research is primarily based in the local context of Cyprus. Therefore, there is a need for further studies exploring EFL teachers' beliefs about AI in pronunciation instruction across diverse educational settings. Such studies would provide a more comprehensive understanding of how contextual factors influence the integration of AI and offer findings that could be generalized to broader, international contexts. Moreover, while the current study sheds light on key beliefs and practices, a more in-depth qualitative approach is needed to explore the underlying reasons for teachers' choices and to examine how various contextual, pedagogical, and personal factors shape their perceptions and use of AI in pronunciation instruction. This deeper qualitative analysis could provide a richer and better understanding of the complexities involved in AI adoption in language teaching, ultimately guiding more effective implementation strategies.

## Conflict of interest

## Acknowledgments

This study was conducted with the support of the Phonetic Lab at the University of Nicosia. I would like to thank all the participants for their valuable contributions to the study, as well as to everyone who assisted with the distribution of the survey.